\documentclass[numbers]{article}

\usepackage[final]{nips_2017}

\usepackage[utf8]{inputenc} 
\usepackage[T1]{fontenc}    
\usepackage{booktabs}       
\usepackage{amsfonts}       
\usepackage{nicefrac}       
\usepackage{microtype}      

\usepackage{adjustbox}
\usepackage{tikz}

\usepackage{caption}
\usepackage{lipsum}

\usepackage{algorithm}
\usepackage{algorithmic}

\usepackage{amsmath}

\usepackage{graphicx}
\usepackage{wrapfig}

\usepackage[colorlinks = true,
            linkcolor = blue,
            urlcolor  = blue,
            citecolor = blue,
            anchorcolor = blue]{hyperref}
            
\usepackage{enumitem}
\usepackage{natbib}
\usepackage{graphicx}

\title{Characterizing Attribution and Fluency Tradeoffs for Retrieval-Augmented Large Language Models}

\author{%
  Renat Aksitov \\
  Google Research \\
  \texttt{raksitov@google.com} \\
  \And
  Chung-Ching Chang \\
  Google Research \\
  \texttt{ccchang@google.com} \\
  \And
  David Reitter \\
  Google Research \\
  \texttt{reitter@google.com} \\
  \And
  Siamak Shakeri \\
  Google Research \\
  \texttt{siamaks@google.com} \\
  \And
  Yunhsuan Sung \\
  Google Research \\
  \texttt{yhsung@google.com} \\
}

\date{February 2023}

\begin{document}

\maketitle

\begin{abstract}
Despite recent progress, it has been difficult to prevent semantic hallucinations in generative Large Language Models. One common solution to this is augmenting LLMs with a retrieval system and making sure that the generated output is attributable to the retrieved information. Given this new added constraint, it is plausible to expect that the overall quality of the output will be affected, for example, in terms of fluency. Can scaling language models help?

Here we examine the relationship between fluency and attribution in LLMs prompted with retrieved evidence in knowledge-heavy dialog settings. Our experiments were implemented with a set of auto-metrics that are aligned with human preferences. They were used to evaluate a large set of generations, produced under varying parameters of LLMs and supplied context.

We show that larger models tend to do much better in both fluency and attribution, and that (naively) using top-k retrieval versus top-1 retrieval improves attribution but hurts fluency. We next propose a recipe that could allow smaller models to both close the gap with larger models and preserve the benefits of top-k retrieval while avoiding its drawbacks.
\end{abstract}

\section{Introduction}
Large language models (LLMs) open the door for many downstream applications requiring text generation, including dialog applications like LaMDA \cite{LaMDA}. While the recent approach of scaling LLMs has significantly reduced the model perplexity and, in turn, enhanced sensibleness (as a proxy for fluency) and specificity \cite{Meena}, these models are still known to hallucinate and provide non-factual/outdated information. To address this issue, one typical approach is to rely on a trusted external knowledge source, such as search engines or retrieval systems, to retrieve relevant evidence, and hope the generated response has knowledge grounded to and attributable to the evidence.

Take LaMDA as an example. After the base model generates an initial response, a separate research model generates search queries and gathers additional information from Google Search. Finally, the research model generates the actual response to users according to all information gathered. Experiments showed that combining LaMDA with Google Search has significantly enhanced response factuality \cite{LaMDA}. Similarly, other work, including REALM \cite{REALM}, RAG \cite{RAG}, RETRO \cite{RETRO}, and \cite{RETRO-ED}, incorporates LLMs with custom retrieval systems and shows that the performance on the downstream knowledge intensive task is significantly improved.

Given that grounding generated responses in retrieved evidence acts as a restriction on the output space, we hypothesize that it could, in principle, lead to less fluent outputs. In other words, there might be a kind of a tradeoff relationship between fluency and attribution of the model’s responses. To verify this hypothesis, we design experiments to measure both metrics for the responses generated by LLMs in various settings that sweep over conditions where LLMs may have different degrees of attribution. Despite the fact that adding a trusted external knowledge source helps, the generation may still not be factual for several reasons:

\begin{itemize}
\item{\textbf{Imperfect retrieval query.} The query to the external knowledge source itself could be a text generation and is also vulnerable to hallucinations. Cascading several LLMs may exacerbate hallucinations.}
\item{\textbf{Imperfect external knowledge source.} The imperfection is measured by precision/recall and AUC curves. When irrelevant information is retrieved and provided, are LLMs intelligent enough to ignore the evidence and say I don’t know. Or does it respond from its own parametric memory instead?}
\item{\textbf{Memorization and grounding.} Does the generated text bind to the memory (model parameters) or the evidence when they are in conflict? Ideally, it is desirable to have a mechanism to control and take precedence between grounding and memorization.}
\item{\textbf{Sampling.} To increase the semantic variability, LLMs typically adopt sampling when decoding response sequences. With higher sampling temperature, the token distribution becomes more even and has a better chance to sample next tokens that are less likely. After rare tokens are sampled, hallucinations could happen, but the LM has no choice and is bound to continue and finish the response sensibly.}
\end{itemize}

 In this work we simplify and eliminate the first two factors by choosing the QReCC dataset where we have access to golden evidence that supports the golden response. The imperfection in the retrieval system is under control and can be simulated. For memorization and grounding, we hypothesize that we can leverage prompting for controlling the knob between them. Emergent abilities \cite{Emerge}, like few-shot learning and chain-of-thoughts reasoning, leverages promptings as context to facilitate new capabilities of the model.
 
 Overall, we are conducting an extensive study, aimed to find out how different model parameters (such as scale or decoding temperature) and the context provided to the model (such as dialog turns, retrieved facts or instructions), can affect fluency and attribution. To be able to better describe the results of our experiments, we introduce the concepts of global and local tradeoff, motivated by prior work. Furthermore, we identify promising ways to reduce or mitigate these tradeoffs.

\section{Related Work}
To the best of our knowledge, there is no literature on the trade-off between fluency and attribution. There are, however, two recent papers, that are looking into different tradeoffs, that we’d like to highlight:
\begin{itemize}
\item{The paper from Zhou et al. \cite{APE} introduces Automatic Prompt Engineer (APE) for automatic instruction generation and selection, and looks into the tradeoff between truthfulness and informativeness. They conclude that memorization is not completely reliable: with being more informative (finer details), the model’s response becomes less factual. They use different (auto-generated) instructions to get different values of a tradeoff. Notably, the model (and, for the most part, the size of the provided context) are fixed.}

\item{On the other hand, the paper from Gao et al. \cite{RARR}, looks into improving factuality through post-hoc research and revision. As part of the experiments, they investigate the tradeoff between attribution and preservation scores, where preservation measures the similarity of the response before and after editing, and the response before editing should be fluent/sensible by the nature of LLM generations. They argue that the better way to look into tradeoffs is through F1 score, given that F1 will be low if either attribution or preservation is low, signifying that both metrics are (equally) important. Notably they also use end-to-end NLI as a proxy for attribution. This paper looks into tradeoffs between 3 different models, EFEC \cite{EFEC}, LaMDA, and RARR.}
\end{itemize}

Note that these 2 papers effectively are using 2 different views on what having a tradeoff means.
\begin{itemize}
    \item[]{\textbf{Global View}: \cite{RARR} examines 3 different models on 3 different datasets and talks about absolute position of aggregated dots within the corresponding 2d plots. Mitigating tradeoff in this sense simply means choosing the best model for the task among the set of available models. We can say that this is a global tradeoff.}
    
    \item[]{\textbf{Local View}: at the same time \cite{APE} takes a fixed model and fixed dataset, limits changes to the instructions in the prompt and optimizes those instructions in order to hit different positions within the 2d plot for auto-metrics. Reducing tradeoff in this sense means choosing the best possible way to perform inference when the model, the task and a certain preference between 2 selected metrics are already decided on. We can say that this is a local tradeoff.}
\end{itemize}

\section{QReCC Dataset}
For the experiments we use a knowledge intensive conversational dataset Question Rewriting in Conversational Context (QReCC, \cite{QRECC}) – an end-to-end open-domain QA dataset consisting of 14K conversations with 81K QA pairs.

Each of 14K conversations consists of a series of questions and answers (a dialog history), followed by the final (not answered yet) query. A golden answer, along with the webpage the answer was extracted from, is also provided for each conversation.

We use a fully decontextualized version of QReCC in our experiments. We decided to use this version after doing manual inspections and discovering that some contextualized conversations are hard to parse even for humans. A side-effect of using a decontextualized version is that for some dialogs the last turn with the user's question has all the required information and the model doesn’t need to know any additional dialog history to produce sensible responses within the full dialog history. Importantly, there are also plenty of dialogs where this is not the case and simply knowing the decontextualized last turn is not enough.

One notable property of using this dataset is that the goal of being sensible within QReCC QA style dialogs is aligned with the goal of being attributable to the evidence. This wouldn’t always be the case and is important for the way the tradeoff behaves. For example, for more chit-chatty datasets the goal of being sensible could in the first place be aligned with being engaging/interesting and mostly independent from the goal of being attributable.

Many dialogs in QReCC dataset assume that the specific Wikipedia article (and, occasionally, even the specific place within the article) is known during the conversation. For example, some turns could refer to “this article” or similar. Such conversations are not desirable for our use case, so we filter them out. Based on manual examinations we further filter out examples where evidence is too long, where history is not well-formed (in some of the examples there are missing turns) or where the golden answer is a single word. See Appendix for the full impact of filtering on the dataset size.

For our experiments we randomly select 100 examples from the remaining after filtering 324 examples in the dev split. We also manually verify these 100 examples to ensure dialog quality.

\section{Human Evals}
\subsection{Pilot}
Meena paper \cite{Meena} introduces a proxy for fluency in the form of Sensibleness and Specificity Average metric (SSA), while paper \cite{AIS}  presents an evaluation framework called Attributable to Identified Sources (AIS) for assessing the output of natural language generation models for attributability. Both SSA and AIS assume the use of human raters and in the ideal world with infinite resources all our experiments would be evaluated this way.
    
To get a sense of the problem, we started with conducting small pilot human eval, 400 examples in total. We sampled 100 dialogs from QReCC, as described above, and generated responses for them using PaLM 540B model \cite{PaLM} with 4 different setups (see next section for the specifics about how we use PaLM’s native dialog prompt):
\begin{enumerate}
\item{temperature = 0.0 and “no evidence”}
\item{temperature = 0.7 and “no evidence”}
\item{temperature = 0.0 and “golden evidence”}
\item{temperature = 0.7 and “golden evidence”}
\end{enumerate}

We have run human SSA eval for all 4 pilot setups and human AIS eval for setups \#3 and \#4. We are assuming here that attributability for setups \#1 and \#2 is 0, given that no evidence is provided to the model (see also a note on definition of “attributable” in  the Appendix).

\subsection{“Merged” Setup for AIS}
Human evals for the pilot 400 examples ended up being fairly expensive, both money and timewise, so we started looking into possible ways to reduce these costs.

As described in the previous section, we use the same 100 dialogs in different setups, which means that for the same example a lot of information is repeated. For example, for the pilot setups \#3 and \#4 evidence and dialog history are the same for the same example, and only the final generated response might differ. This suggests that an eval optimization could be warranted, where instead of doing first all 100 examples from \#3 and then all 100 examples from \#4, we can do them in parallel. That is, the rater is asked to rate the same example first for \#3 and then immediately for \#4, so that he doesn’t need to re-read evidence and dialog history for this example.

\subsection{Alignment of “Merged” vs “Separate”}
   What we discovered is that an implicit SxS of “merged” evals systematically changes human ratings. More specifically, we looked into the following parts of AIS evaluation:
\begin{enumerate}
    \item{"understand",}
    \item{"relevant",}
    \item{"consistent",}
    \item{"evidence relevant",}
    \item{"evidence contradicts",}
    \item{"evidence supports",}
    \item{"attributable".}
\end{enumerate}

All of them are binary classification questions, which allows us to easily define a “single number” aggregated metrics for each. We simply do a majority voting (we use 5 raters, so always well-defined) for each example and then take a mean over the whole set of 100 examples.

    We can now compute these aggregated metrics for the 7 questions in the following 4 scenarios:
\begin{enumerate}[label=\Alph*]
\item{T=0.0 for generation, “separate” evaluation}
\item{T=0.0 for generation, “merged” evaluation}
\item{T=0.7 for generation, “separate” evaluation}
\item{T=0.7 for generation, “merged” evaluation}
\end{enumerate}
Our hope/expectation was that the metrics from (A) and (B) pair will be similar, as well as from (C) and (D) pair. 

Instead we have discovered that questions (2), (4) and (6) are experiencing systematic shift under “merged” setup: (B) is “better” than (A), and (D) is “worse” than (C). In other words under “merged” eval the raters prefer outputs generated with T=0.0 more and with T=0.7 less. It’s not immediately clear whether “merged” ratings are better than “separate” (we do see examples in both directions), but they are significantly different at least for a subset of AIS questions. 

We will use the difference in 2 eval setups to estimate human’s “uncertainty rate” on our data. Specifically, we will assume “separate” evals as ground truth and measure accuracy of “merged” evals against it.

\section{Auto-Metrics}
\subsection{End-to-end NLI for Auto-AIS}
We further look into end-to-end NLI in various flavors as a proxy for “attributable”:
\begin{itemize}
\item{Flavor v1: \{Golden Evidence, Question\} => \{Answer\}}
\item{Flavor v2: \{Golden Evidence\} => \{Question, Answer\}}
\item{Flavor v3: \{Golden Evidence, Question\} => \{Question, Answer\}}
\end{itemize}

For example, "v2" concatenates the question that immediately precedes the model’s answer in the dialog with the answer and checks if there is entailment between golden evidence and the concatenated pair. Out of these "v3" tends to work better, as measured by alignment with human ratings. But even “v3” only gets 75\% accuracy against human AIS, which improves over simply predicting the majority label (60\%), but not by much.

RARR \cite{RARR} shows a boost in AIS performance from splitting the (long) answers into individual sentences, and running NLI for each one. In our case answers are typically short, but evidence is long and could be up to 900 tokens in some cases. We hypothesize that this is the source of the problem and, to verify, we split the evidence by sentence boundaries and compute all NLI scores against fixed size sliding windows of K sentences, then take the maximum of these “localized” scores as the final score. Similar to \cite{SummaC}, we find that granularity of K=2 performs best and boosts accuracy against humans to 87\%. At the same time, human’s uncertainty rate, estimated from “merged” against “separate” AIS, stands at 93\%. In other words, “localized” NLI almost closes the gap with human evaluations and can be considered a reliable proxy.

\subsection{PaLM for Auto-SSA}
We have created custom few-shot prompts for PaLM to enable the model to output SSA ratings (we did build prompts for both Sensibleness and Specificity, but ended up actively using only the prompt for Sensibleness in the experiments). 

\begin{wraptable}{r}{8cm}
\centering
\setlength{\arrayrulewidth}{0.5mm}
\begin{tabular}{|c | c | c | c | c |} 
 \hline
 Pilot slice & MSE & 1 -> 0 & 0 -> 1 & Acc \\
 \hline
 t=0.0, no evidence & 0.033 & 4\% & 1\% & 5\% \\
 t=0.0, golden evidence & 0.016 & 1\% & 1\% & 2\% \\
 t=0.7, no evidence & 0.040 & 4\% & 1\% & 5\% \\
 t=0.7, golden evidence & 0.026 & 2\% & 4\% & 6\% \\
 \hline
\end{tabular}
\captionsetup{format=plain, font=large, labelfont=bf}
\caption{sensibleness prompt alignment}
\label{table:palm_sens}
\end{wraptable}

The prompts are built based on the human ratings from setup 2 in the pilot data (i.e. temperature = 0.7 and “no evidence”). We use linear search in the example space with the goal to predict a score from the “aggregated rater”. In other words, if 2 out of 5 raters said “Yes, sensible” and the remaining 3 said “No”, we will assign a sensibility score of 0.4 to this specific example. We use the remaining 3 pilot slices as a validation set for the resulting prompt (table \ref{table:palm_sens}):

The final PaLM prompt for Sensibleness has an error rate versus human raters of $\approx 4\%$ on average when measured on the validation sets (row 1, 2 and 4 in the table \ref{table:palm_sens}). The fragment of the final prompt (see Appendix for a complete version):

\fbox{\begin{minipage}{50em}
\textcolor{violet}{\textbf{Instructions}}: Does B’s final reply in the dialog below make sense to you? Use your common sense here. Is the response completely reasonable in context? Then rate it as '1.0'. If anything seems off — confusing, illogical, out of context, lacks common sense — then reduce the rating accordingly. Slightly illogical? '0.8'. Complete nonsense out of context? '0.0'
\bigbreak
\textcolor{violet}{\#\#\#} \\
\textcolor{violet}{\textbf{Dialog}}: \\
\textcolor{violet}{\textbf{A}}: who celebrates new year first in the world? \\
\textcolor{violet}{\textbf{B}}: Tonga and Kiritimati, part of Kiribati, are examples of the first places to welcome the New Year \\
\textcolor{violet}{\textbf{A}}: who celebrate new year last in the world? \\

\textcolor{violet}{\textbf{Final reply}}: \\
\textcolor{violet}{\textbf{B}}: Samoa and American Samoa are the last places to welcome the New Year, as they are the first to see the sunrise on January 1st. \\
\textcolor{violet}{\#\#\#}
\bigbreak
\textcolor{violet}{\textbf{Answer}}: 0.4
\bigbreak
...
\end{minipage}}

\section{Experimental Setup}
\subsection{Native Dialog Prompt}
To support conversational datasets with PaLM, we adopt “native” dialog prompts for PaLM, based on how dialog data was passed to the model during the training. An example of the format is shown below (notice that the example has 3 participants):

\fbox{\begin{minipage}{20em}
\textcolor{olive}{0 -1 0} Knock Knock \textcolor{olive}{[eot]} \\
\textcolor{olive}{1 0 1} Who's there? \textcolor{olive}{[eot]} \\
\textcolor{olive}{2 1 0} Interrupting cow \textcolor{olive}{[eot]} \\
\textcolor{olive}{3 1 2} Nobel \textcolor{olive}{[eot]} \\
\textcolor{olive}{4 3 1} Nobel who? \textcolor{olive}{[eot]} \\
\textcolor{olive}{5 4 2} That's why I knocked \textcolor{olive}{[eot]} \\
\textcolor{olive}{6 5 1} <PaLM to complete>
\end{minipage}}

Here, the first number is a turn’s index, the second is the index of a parent’s turn and the third is a speaker’s id. Note that this format supports multiple speakers and non-linear structure of the conversations.
    
The use of native dialog prompts for the response generation has an indirect effect of forcing the model to stay sensible within the dialog history. It also simplifies parsing by allowing us to simply stop at the next [eot] token.

\subsection{Advanced Promptings}

Emergent abilities \cite{Emerge}, like few-shot learning and chain-of-thought reasoning, leverage promptings to facilitate new model capabilities. With advanced promptings, we aim to make a sensible model (the model optimized for perplexity) more attributable. In other words, by structuring prompt with “instructions”, “facts” and “dialog history”, we hope the language model will generate a (sensible) response attributable to the given evidence.

The full prompt might look something like this:

\fbox{\begin{minipage}{50em}
\textcolor{olive}{\textbf{Instructions}}: use the information from the provided “fact” to answer the question \\

\textcolor{olive}{\textbf{Fact}}: Racing career [ edit ] Early racing career [ edit ] Kulwicki began his racing career as a 13-year-old kart racer. [10] His father built engines as the crew chief for Norm Nelson and Roger McCluskey 's United States Automobile Club (USAC) racecars. [1] [12] Because his work involved travel, Kulwicki's father was unable to help his son at most kart races, [9] ... (truncated) \\

\textcolor{olive}{0 -1 0} When did Alan Kulwicki start racing? \textcolor{olive}{[eot]} \\
\textcolor{olive}{1 0 1} Kulwicki began his racing career as a 13-year-old kart racer. \textcolor{olive}{[eot]} \\
\textcolor{olive}{2 1 0} Was Alan Kulwicki able to race cars at the young age of 13? \textcolor{olive}{[eot]} \\
\textcolor{olive}{3 0 1}
\end{minipage}}

To better understand the impact of various components, we further adjust structure of the prompts along several dimensions:
\begin{itemize}
\item{“instructions” could be either present or absent,}
\item{similarly, dialog history could be also present or absent,}
\item{and the provided evidence could be:
\begin{itemize}
\item[]{golden,}
\item[]{retrieved,}
\item[]{absent,}
\item[]{non-evidence (i.e. guaranteed not to be golden evidence).}
\end{itemize}}
\end{itemize}

For the retrieved evidence we use BM25 and look into top-1, top-2 and top-3 retrieval (e.g. in the latter case, three facts will be provided in the prompt rather than one). Also check the notes about “simulated” retrieval in the “results” section.

We arrive into a “full grid” of experiments by running the prompting setups described above in 6 settings each: 3 different model sizes (8B, 62B and 540B) and 2 sampling temperatures (0.0 and 0.7).

As was mentioned in the previous section, conducting the full grid of experiments with human raters is unrealistic. Instead, after confirming that the auto-metrics are well-aligned with the human ratings from the pilot, we apply Auto-AIS and Auto-SSA to the full grid of experiments for further analysis.

\section{Results and Discussion}
In this part we will examine the auto metrics on a full grid of experiments and summarize the findings in several takeaways about global and local tradeoffs. The overall plan is as follows.

First, we will find out that the parameters of the context structure, like presence or absence of the evidence or dialog history, have a large scale effect on the value of the global tradeoff as measured by F1 (takeaway 1), while varying the model parameters (i.e., size or sampling temperature), has only a medium scale effect (takeaway 2).

Next, we will look into quantifying a local tradeoff and conclude that it is inherently connected to having some additional, “hidden” constraints, like context size (takeaway 3). Based on this observation, we will design a set of synthetic experiments to demonstrate how local tradeoff manifests for top-1 retrieval (takeaway 4) under restricted context ratio “budget”, and will use the same set of experiments to argue further that using top-k retrieval for higher recall leads to improved attribution at a cost of reduced fluency (takeaway 5). We will also look into the effect of input-level re-ranking on both global and local tradeoffs (takeaway 6).

Finally, we will discuss how the results / takeaways affect our understanding of global and local tradeoffs and propose a way to combine all the takeaways together into a recipe in which a small (8B) model is used with top-k retrieval and re-ranking to achieve values of global tradeoff / F1 that are comparable to those produced by a large (540B) model with oracle knowledge.

\subsection{Four Clusters}
\begin{figure}[!htbp]
    \centering
      \includegraphics[width=0.75\textwidth, keepaspectratio]{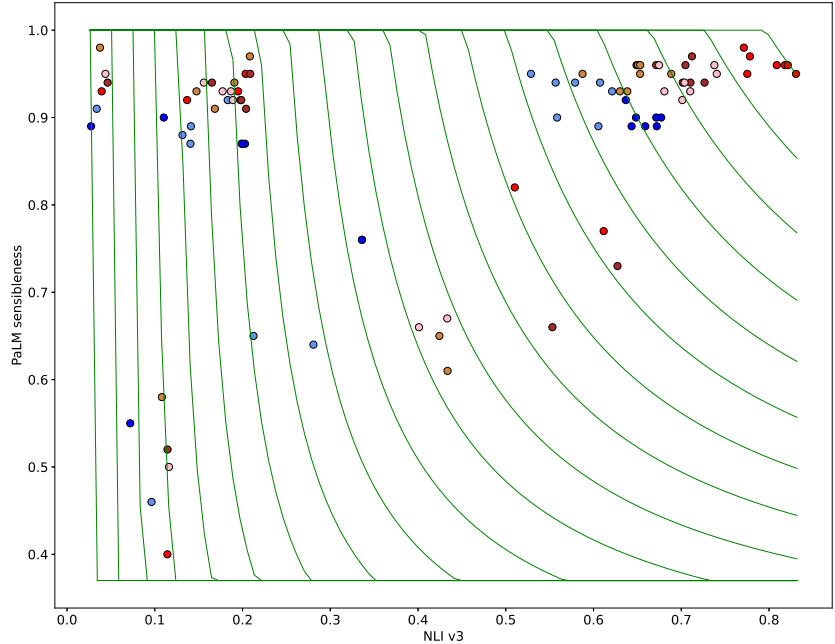}
      \captionsetup{format=plain, font=footnotesize, labelfont=bf}
      \caption{Full grid, 4 clusters}
      \label{fig:clusters}
\end{figure}

Let’s start with a full grid plot. Each dot in the figure \ref{fig:clusters} corresponds to aggregated metrics for selected 100 examples where the final response is generated by the same model:
\begin{itemize}
\item{Different colors mean 3 different sizes of the model - blue for small (PaLM, 8B), brown for medium (PaLM, 62B) and red for large (PaLM, 540B),}
\item{The light version of the particular color (light blue, light brown, pink) means temperature=0.7 (vs temperature=0.0 otherwise),}
\item{Green curves are iso-F1 levels.}
\end{itemize}

There are 4 distinct clusters in this figure as outlined in the table \ref{table:clusters}:
\begin{table}[!htbp]
\centering
\setlength{\arrayrulewidth}{0.5mm}
\begin{tabular}{|c | c | c |} 
 \hline
   & Dialog history present & Dialog history absent \\
 \hline
   Golden evidence present & Top right & Center \\
 \hline
   Golden evidence absent & Top left & Bottom left \\
 \hline
\end{tabular}
\captionsetup{format=plain, font=large, labelfont=bf}
\caption{Cluster types}
\label{table:clusters}
\end{table}

Should we expect the “center” cluster to be in the bottom right instead? Note that “dialog history absent” means that the last dialog turn (i.e. user’s question) is still provided, which in our case of decontextualized queries often has enough information for the model to keep the dialog sensible, even without knowing the rest of it.

\textbf{Takeaway 1}. Presence or absence of the evidence and/or dialog history have a \textit{large scale} effect on the value of the tradeoff (where tradeoff is considered in accordance with \textit{global} view).

\subsection{Golden Evidence}
\begin{figure}[!htbp]
    \centering
      \includegraphics[width=0.75\textwidth, keepaspectratio]{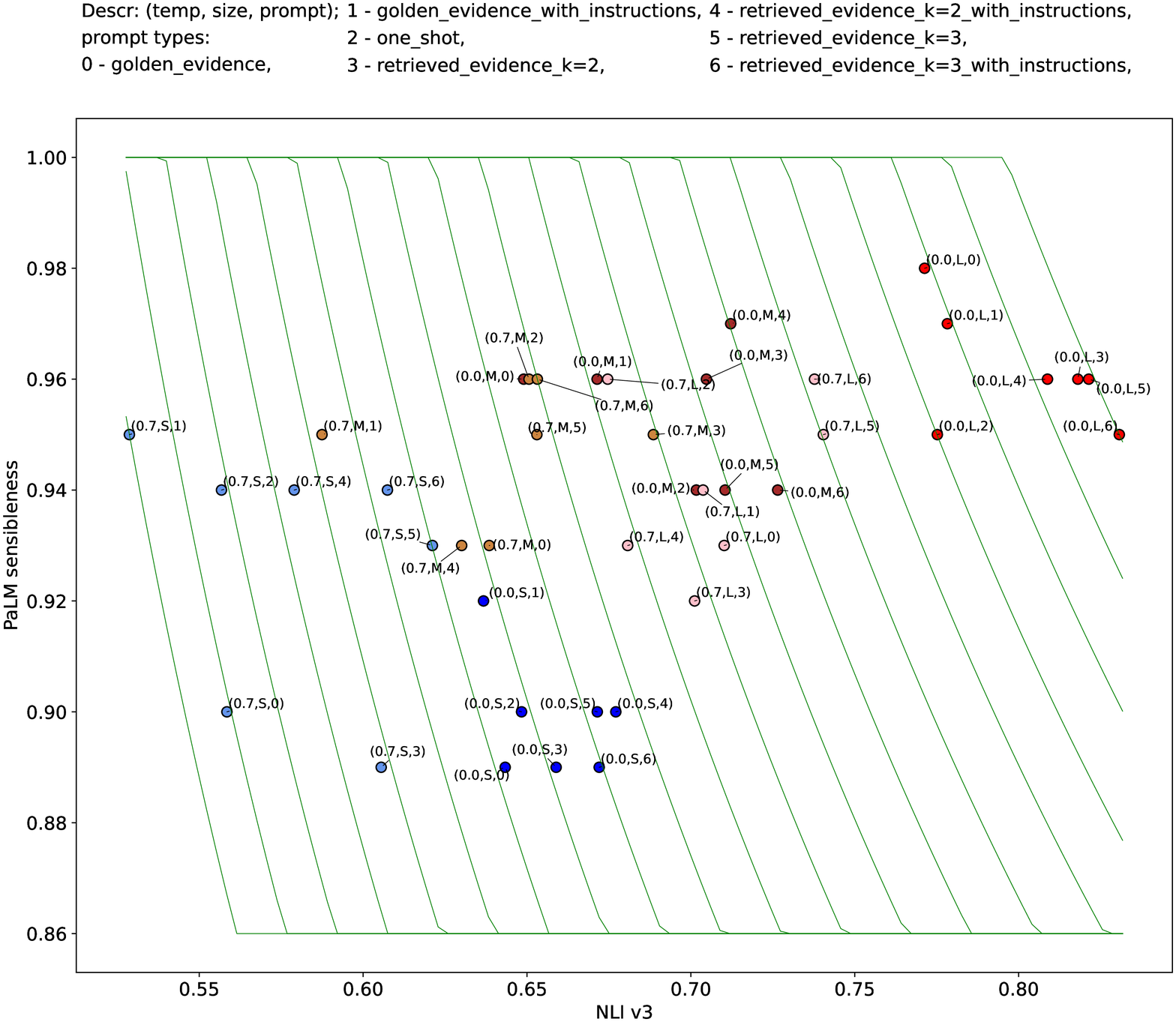}
      \captionsetup{format=plain, font=footnotesize, labelfont=bf}
      \caption{Cluster with golden evidence}
      \label{fig:golden}
\end{figure}

Next, let's zoom-in to the “top right” cluster (figure \ref{fig:golden}), i.e. all the models for which both golden evidence and a full conversation history are available.

Here we have 4 well-defined F1 bands (from left to right, in the direction of increasing F1):
\begin{itemize}
\item{\textit{light blue} (small models, t=0.7)}
\item{\textit{blue} (small models, t=0.0) and \textit{light brown} (medium models, t=0.7)}
\item{\textit{brown} (medium models, t=0.0) and \textit{pink} (large models, t=0.7)}
\item{\textit{red} (large models, t=0.0)}
\end{itemize}

In other words, for our selection of models, sampling temperature and model size have comparable effect on the value of the global tradeoff and one can choose between smaller model in greedy mode or larger model with higher temperature and expect to end up within the same F1 band.

\textbf{Takeaway 2}. Reducing temperature and increasing model size have a comparable \textit{medium scale} effect on the F1.

\subsection{Large Model}
Let’s zoom-in more and look into the large models only:

\begin{figure}[!htbp]
    \centering
      \includegraphics[width=0.75\textwidth, keepaspectratio]{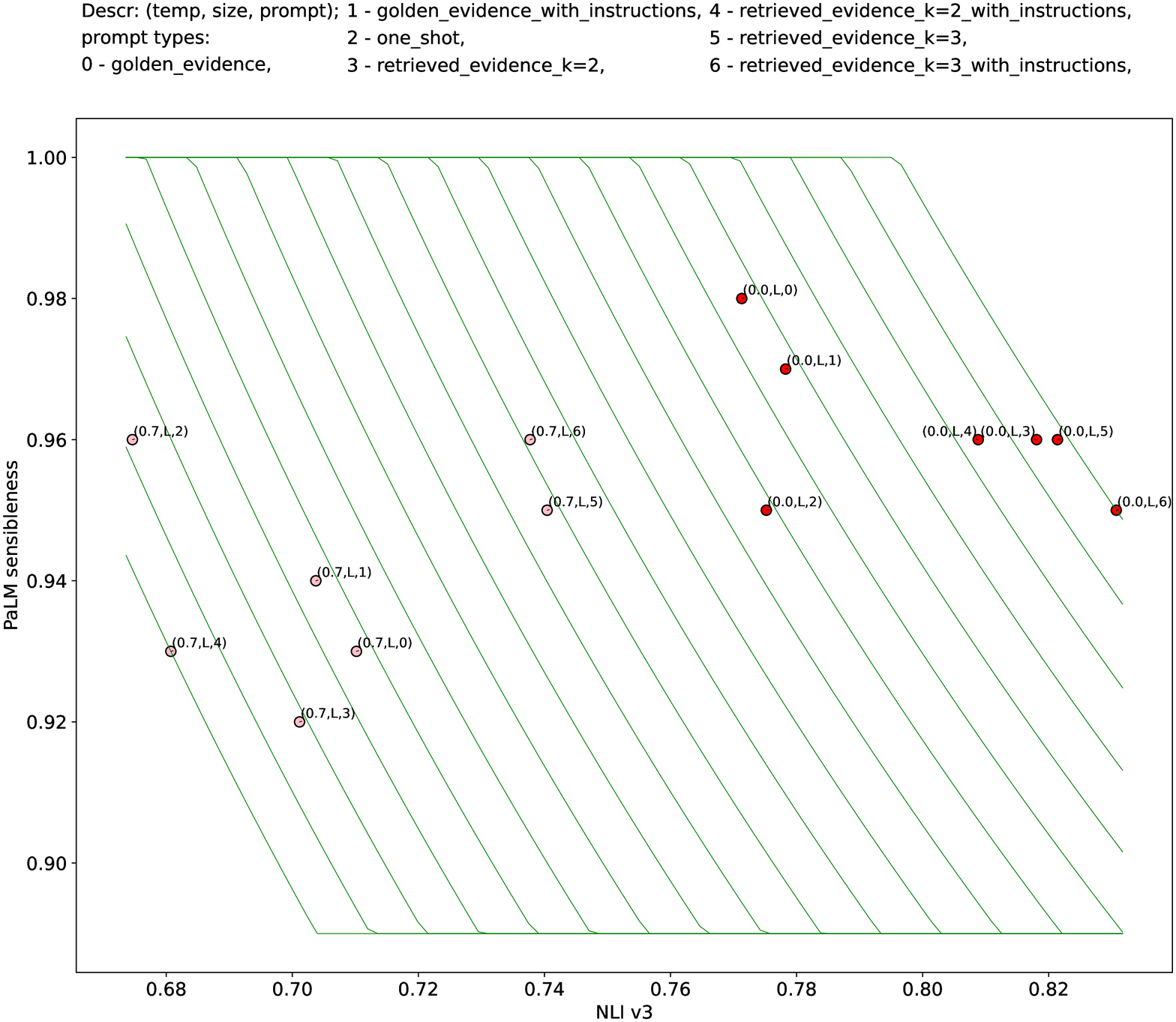}
      \captionsetup{format=plain, font=footnotesize, labelfont=bf}
      \caption{Large models with golden evidence}
      \label{fig:large}
\end{figure}

High-temperature experiments on the left (i.e. where 1st value in the descriptor tuple is t=0.7) are well-separated from the low-temperature ones (t=0.0) on the right by F1 bands, as we have already seen in the previous figure for the medium scale effects. The rest of the variance within the sets of same-temperature experiments comes from the differences in the prompt: presence or absence of instructions, additional retrieved evidence besides gold (i.e. top-2 or top-3).

Can we start looking for tradeoffs in the local sense for these sets? In other words, is it enough to fix model size and sampling temperature (and, implicitly, the dataset) to see systematic local tradeoff for different prompt structures (i.e. within same-color experiments in figure 3)? For the most part, the answer is “no”.

The prompts that are somewhat comparable are the prompts that differ only in “instructions” being present or not. Specifically “1” vs “6”, “0” vs “4” or “3” vs “5” for prompt type (3rd value in the descriptor tuple).

Why are the rest not comparable? We hypothesize that the missing variable is the size of the context (i.e. the combined size of everything that was passed to the model as input: dialog history, one or more retrieved facts, instructions, etc). For an extreme example, consider the prompt with “no dialog history” and “no evidence” and compare it to the prompt with “full dialog history”, “instructions” and “3 retrieved facts”. The model that is using the former is never going to be more fluent than the latter purely because it has more degrees of freedom.

\textbf{Takeaway 3}. Comparing experiments with the context of similar size is important for quantifying the \textit{local} tradeoff.

\subsection{Restricted Context and Retrieval}
PaLM is able to effectively handle fairly large context ($\approx 2000$ sentence pieces), so to investigate the impact of limited context size we will restrict it artificially in the following way:
\begin{itemize}
\item{Let’s start (in the top left corner) with all dialog turns (2 * n + 1 turns) and an empty evidence,}
\item{Keep dropping turns one by one from the dialog (from the top, i.e. starting from the older turns),}
\item{Simultaneously, let’s add sentences to the evidence (from the beginning) in such a way that the total amount of sentence pieces in dialog and evidence stays approximately the same (more specifically, the ratio of [the remaining dialog turns to the full dialog history] plus [the ratio of the newly added evidence to the full evidence] is kept close to 1),}
\item{We end (in the bottom right corner) with no dialog turns (unlike with a “center” cluster before, the end state does not even have a user’s query) and full evidence.}
\end{itemize}

The figure \ref{fig:top-1} below shows this process for the 540B model with temperature=0. Green dots mean that “non-evidence” was used, the corresponding blue dots use “golden” evidence, and the connecting “gray” lines could be treated as “top-1 evidence” for various values of recall (with “green” end being $recall@1 = 0\%$ and “blue” end being $recall@1 =100\%$).

\begin{figure}[!htbp]
    \centering
      \includegraphics[width=0.75\textwidth, keepaspectratio]{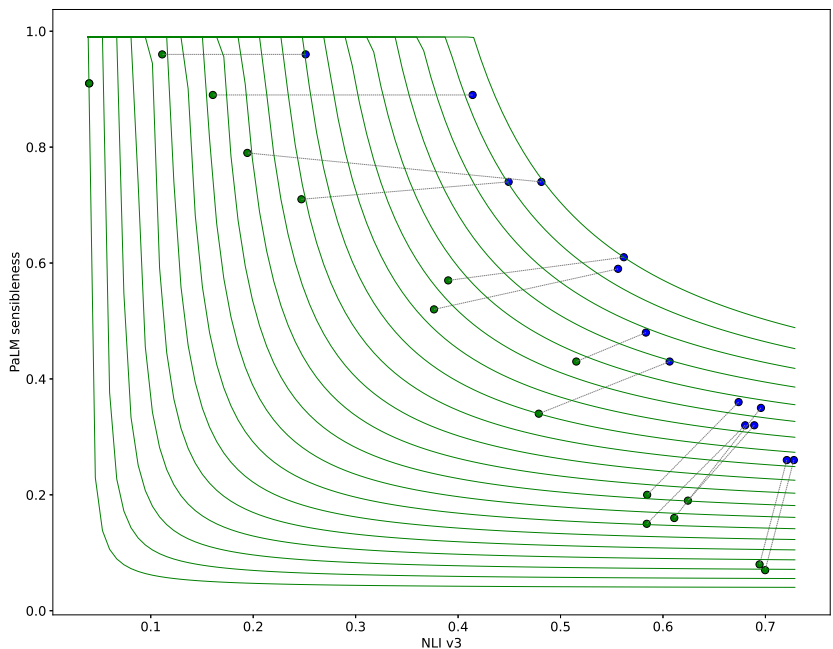}
      \captionsetup{format=plain, font=footnotesize, labelfont=bf}
      \caption{Restricted context and top-1 retrieval}
      \label{fig:top-1}
\end{figure}

More specifically, we first compute the values of auto-metrics for the same 100 examples both with “non-evidence” and with “golden” evidence. Those values are then used to obtain auto-metrics of the “average” retrieval system with a given value of $recall@1 = X\%$ for various values of X from 0 to 100. These interpolated values give us “gray” lines of simulated top-1 retrieval.

Notice that curves defined by blue and green dots respectively are (mostly) well-aligned with iso-F1 lines.

\textbf{Takeaway 4}. There is a clear \textit{local} tradeoff for models with fixed context ratio between dialog and evidence.

There are several plausible ways to re-interpret this simplistic synthetic experiment in more practical terms. 

 One interpretation is to think about a two-stage retrieval system, where first satge retrieves (long) documents and then second stage selects relevant parts within them with some kind of reading comprehension system. More / less space allocated for evidence in the synthetic setup will then correspond to having a better / worse second stage in a “real” two-stage setup.

Another interpretation could be in terms of top-1 vs top-k retrieval. In this case adding additional pieces to the evidence in our artificial setup corresponds to switching from retrieving $X$ items to retrieving $X + 1$.

The latter interpretation leads us to the conclusion that in the case of limited context “budget” it’s not necessarily advantageous to use top-k evidence for improved recall. 

\begin{figure}[!htbp]
    \centering
      \includegraphics[width=0.75\textwidth, keepaspectratio]{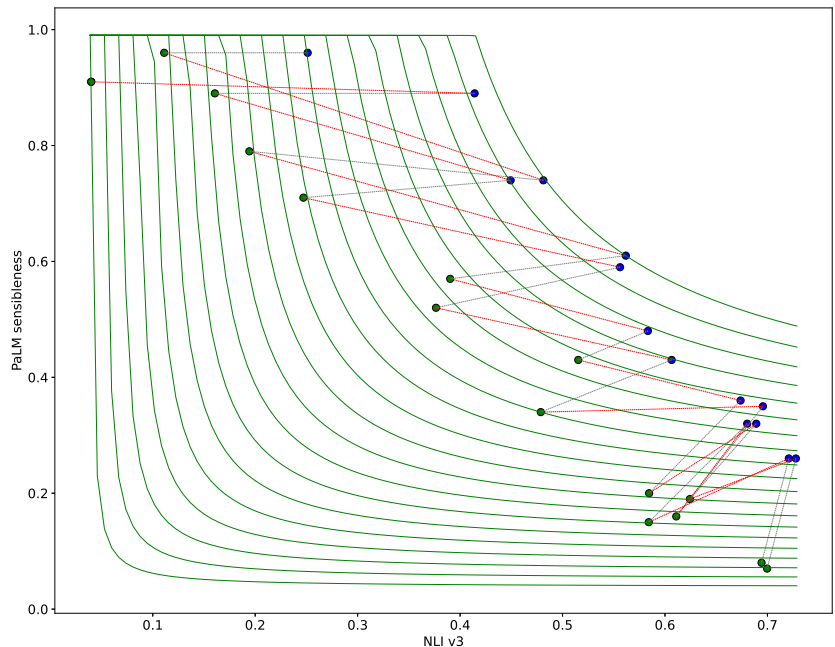}
      \captionsetup{format=plain, font=footnotesize, labelfont=bf}
      \caption{Restricted context and top-k retrieval}
      \label{fig:top-k}
\end{figure}

For example, in Figure \ref{fig:top-k}, top-k evidence will lead us to lower (in sensibleness) blue dots compared to dots for top-1 evidence and, in general, we might be losing fluency while gaining attributability (check the slope of the “red” lines below for approximation of the effect that using top-3 evidence could have). I.e. while we are moving faster along the retrieval lines when using top-3 retrieval (due to recall@3 > recall@1), we are moving in a different direction (along “red” lines instead of “gray”).

\textbf{Takeaway 5}. Improved recall from top-k retrieval can often lead to increased attribution and reduced fluency, compared to top-1 retrieval.

\subsection{Re-Ranking or Input-Level Ensembling}
Let’s now look into local tradeoff from a different angle, which should allow us to connect global and local tradeoffs together, and provide a way to mitigate tradeoff to a degree.

Consider 7 “red” experiments from figure 3 (all with golden evidence). They consist of the same 100 input examples with 7 different prompt structures, which could further be re-ranked on the input example level by auto-metrics to produce “aggregated” experiments. I.e. we can choose 1 out of 7 examples, based on some selection criteria and do it this way for all 100 input examples to build an “aggregated” or re-ranked experiment.

\begin{figure}[!htbp]
    \centering
      \includegraphics[width=0.75\textwidth, keepaspectratio]{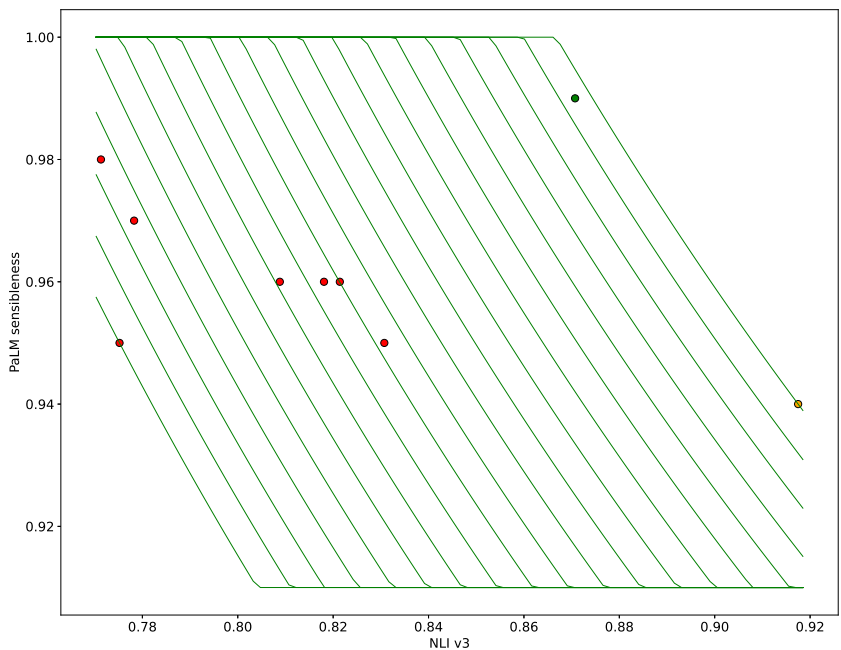}
      \captionsetup{format=plain, font=footnotesize, labelfont=bf}
      \caption{Aggregating 540B models by NLI (orange dot) and by Sensibleness, then NLI (green)}
      \label{fig:re_rank_large}
\end{figure}

For example, if we want to improve an attribution score, we can choose to simply maximize NLI as our selection criteria above. This is represented by an “orange” experiment in figure \ref{fig:re_rank_large}. Alternatively, we can first filter out examples that were deemed non-sensible (i.e. PaLM produced a score less than 0.5) and only then chose those that maximize NLI. This way we will arrive at a “green” experiment instead.

\begin{figure}[!htbp]
    \centering
      \includegraphics[width=0.75\textwidth, keepaspectratio]{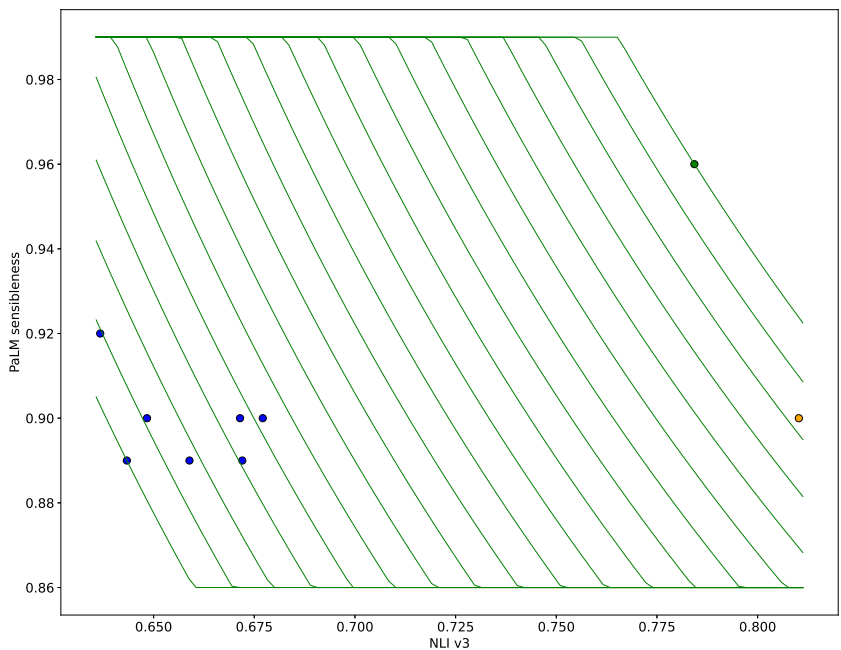}
      \captionsetup{format=plain, font=footnotesize, labelfont=bf}
      \caption{Aggregating 8B models by NLI (orange dot) and by Sensibleness, then NLI (green)}
      \label{fig:re_rank_small}
\end{figure}

Notice that while original “red” experiments are not comparable for local tradeoff purposes, as was explained before, “green” and “orange” experiments are produced under the same constraints and, as a result, could be considered comparable. These two aggregated experiments demonstrate clear local tradeoff between fluency and attribution, and a significant improvement in F1 from the initial non-aggregated experiments. Figure \ref{fig:re_rank_small} shows the same re-ranking process applied to the outputs of the small model (notably, the “green” aggregated dot from figure \ref{fig:re_rank_small} falls inside the cluster of “red” unaggregated dots from figure \ref{fig:re_rank_large}).

To summarize this section, re-ranking of the multiple outputs produced by the model with the same size and temperature leads to medium scale improvements in F1, comparable in our experiments to the effect from simultaneously reducing temperature and increasing model size (i.e. approximately doubling the effect from takeaway 2 before) or keeping temperature the same and increasing model size by two orders of magnitude (i.e. going from 8B -> 540B). Compare also to the similar results from \cite{P-R}.

\textbf{Takeaway 6}. Performing multiple inferences with small model (8B) and then re-ranking them by auto-metrics allows to get F1 values of global tradeoff comparable to the ones produced from large model (540B) without re-ranking.

\subsection{Discussion}
Let’s revisit global and local tradeoff based on the results / takeaways. On a high level we can point out the following:
\begin{itemize}
\item{Global tradeoff in general leads to relatively large “moves” that are orthogonal to iso-F1 curves. For example, on figure \ref{fig:clusters} it’s a move from models with “query only” to models with some context and then to the models with a full context. On figure \ref{fig:golden} it’s a movement from left to right corresponding to the model size and temperature changes and on figure \ref{fig:re_rank_large} it’s a move from original “red” dots to the re-ranked “orange” and “green” ones.}
\item{At the same time, local tradeoff typically results in movement alongside the iso-F1 levels (in other words local tradeoff tends to preserve F1 value). For example, on figure \ref{fig:large} it’s small scale moves between the same model where the only difference in the prompt structure is absence or presence of “instructions” ([0.0, L, 6] -> [0.0, L, 1] or [0.7, L, 6] -> [0.0, L, 1]), on figure \ref{fig:top-1} it’s the movement along the curves corresponding to the dots of the same color (blue or green) and on figure \ref{fig:re_rank_large} and \ref{fig:re_rank_small} it’s a move between “orange” aggregated experiment and the “green” one.}
\end{itemize}

Notice further that for local tradeoff to happen, we need to have some kind of constraints being defined. In other words, operating in fully unconstrained mode is not very interesting: we simply choose the largest available model, provide it with a full dialog history and golden evidence, try all kind of different prompt structures and use human raters to cherry pick the best generations from a very large number of inferences.

The constraints could include fixing the overall context size, as done in takeaway 3 or restricting the overall ratio that could be “spent” on dialog and evidence, as done in synthetic experiments from takeaways 4 and 5. The constraint in takeaway 6 is a bit more subtle - the context size here is not restricted, as different “red” experiments have (very) different prompt structures and vary a lot in terms of the overall context size. What is restricted, however, is the number of inferences (7 in this case), the set of prompts (also 7) and the space of functions used for re-ranking (in our case, even though our PaLM-based auto-sensibleness produces pseudo-scores, we have only justified its alignment with human preferences as a boolean metric; if, instead, we’ve build an auto-metric that gives sensibleness as an actual score from [0, 1], we could’ve expanded the re-ranking functions space to, for example, all linear combinations of a * auto-sensibleness + b * auto-attribution). 

Other constraints that could potentially (we haven’t covered those with comprehensive experiments and consider them as promising directions for future work) result in local tradeoff include:
\begin{itemize}
\item{Relative positioning within the context (e.g. putting evidence early in the prompt structure vs putting it close to the end, etc; especially relevant to the case when native dialog prompt format is used, as there could only be single last turn in the dialog and one can expect the last turn in the dialog to have the largest impact on the next turn).}
\item{The number, the specific ratio of the examples and overall cost / efforts spent on the fine-tuning mixture, if one is used (e.g. fine-tuning mixture could have more diversity of examples how to be attributable in different situations, but not enough examples of staying sensible in diverse situations; this could be less relevant for QReCC but more important for chit-chatty datasets we discussed previously).}
\item{Function space constraints for decoding strategy. So far, we have only looked into greedy and temperature-based decoding, but significantly more sophisticated decoding schemes could, in principle, be used and lead to better results (see \cite{fact-sampling} for example) in terms of the F1 value / global tradeoff. We would still expect to see local tradeoff being present within such extended decoding space.}
\end{itemize}

\subsection{Putting Things Together}
Let’s assume that we can afford running (at scale) only the smallest (i.e. 8B) model, but we want to get results that are comparable in F1 value / global tradeoff to large models that are using oracle knowledge of golden passages. Our qualitative experiments suggest the following high-level recipe:
\begin{enumerate}
\item{Choose a retrieval system (for example, Google Search) and value $K_1$, such that recall@$K_1$ is sufficiently close to 1 for the anticipated use case.}
\item{Find a value $K_2 <= K_1$, such that it’s possible to fit $K_2$ retrieved items of the average length into the prompt without exceeding context size (or context ratio) constraints.}
\item{Build auto-metrics, aligned with human preferences for the expected data. If using a large model for Auto-SSA, as we did, to be practical this might require a 2-step approach of first tuning the prompt with a large model and then distilling it into a small model over a large amount of unlabeled data.}
\item{Perform a single run of the chosen retrieval system for each data point (i.e. an input dialog for which we want to generate knowledge-grounded next turn) and retrieve $K_1$ items.}
\item{For the same datapoint, run multiple inferences using the 8B model with advanced promptings as described previously for different values of retrieved items $K \in [1, K_2]$. For each K, run at least ceiling ($K_1$ / K) inferences, to cover all $K_1$ items within prompts that have exactly K retrieved items in them.}
\item{Use Auto-Fluency as a boolean metric followed by Auto-Attribution to re-rank all the inferences for the given datapoint and produce a single “aggregated” inference.}
\end{enumerate}

\section{Conclusion}
We examined the relationship between fluency and attribution for large language models on a factuality-focused dialog dataset (QReCC). We started with human evaluation on a small subset of the data, then used these ratings to align a set of auto-metrics with human preferences from the pilot. We next applied the auto-metrics to a much larger set of experiments that explore a wide range of parameters, both model and context-specific.

To describe the results of our experiments, we have followed implicit definitions from prior work to explicitly introduce and later refine the concepts of global and local tradeoff. We summarized our findings as several takeaways about 2 views of the tradeoff and proposed a way to combine all the takeaways together into a retrieval-augmentation recipe.

\section{Appendix}
\subsection{QReCC Filtering}

\begin{wraptable}{r}{11cm}
\centering
\setlength{\arrayrulewidth}{0.5mm}
\begin{tabular}{|c | c | c | c | c |} 
 \hline
 Filter & dev, \# examples & dev, \% & train, \# examples & train, \% \\
 \hline
 
None & 5698 & 100 & 51108 & 100 \\
no history & 4768 & 83.6 & 42735 & 83.6 \\
even \# of turns & 4172 & 73.2 & 37668 & 73.7 \\
question after question & 4056 & 71.1 & 36584 & 71.6 \\
1 word golden answer & 4033 & 70.8 & 36252 & 70.9 \\
facts are >= 300 tokens & 2869 & 50.4 & 25262 & 49.4 \\
underspecified questions & 2823 & 49.5 & 25118 & 49.1 \\
last turn mentions ‘article’ & 2743 & 48.1 & 24375 & 47.6 \\
exact match in evidence & 324 & 5.7 & 2829 & 5.5 \\
 \hline
\end{tabular}
\captionsetup{format=plain, font=large, labelfont=bf}
\caption{QReCC filtering}
\label{table:data_filtering}
\end{wraptable}

The full list of filters and their impact on the dataset size is listed in the table \ref{table:data_filtering}:

\subsection{Note on Definition of “Attributable”}
For pilot data setups 1 and 2 (i.e. with “no evidence”) we assume that AIS is 0. This, in principle, is a choice that depends on the selected definition of what “attributable” means; some possible alternatives could be:
\begin{itemize}
\item{To always measure attributability against “golden evidence”, independently of what kind of evidence is provided to the model,}
\item{To first retrieve evidence based on the input dialog and generated response (from the indexed corpus of evidence) and measure attributability against this retrieved evidence.}
\end{itemize}

\subsection{Full List of Prompt Types}
Below is the full list of prompt types that we have used in experiments:
\begin{itemize}
\item{no evidence}
\item{with golden evidence}
\item{with golden evidence and instructions}
\item{retrieved evidence, k={2, 3} (“retrieved evidence” here and below always includes golden)}
\item{retrieved evidence and instructions, k={2, 3}}
\item{one-shot (this prompt always uses golden evidence)}
\item{no history, no evidence}
\item{with random non-evidence (guaranteed to not include golden evidence)}
\item{with next best non-evidence (guaranteed to not include golden evidence)}
\end{itemize}

For retrieved evidence, we simply use either BM25 or random selection for retrieval. Given that we can simulate retrieval systems with any given values of recalls, the specific choice of the retrieval system here is not that important.

\subsection{Final Sensibleness Prompt}

\fbox{\begin{minipage}{50em}
\textcolor{violet}{\textbf{Instructions}}: Does B’s final reply in the dialog below make sense to you? Use your common sense here. Is the response completely reasonable in context? Then rate it as '1.0'. If anything seems off — confusing, illogical, out of context, lacks common sense — then reduce the rating accordingly. Slightly illogical? '0.8'. Complete nonsense out of context? '0.0'
\bigbreak
\textcolor{violet}{\#\#\#} \\
\textcolor{violet}{\textbf{Dialog}}: \\
\textcolor{violet}{\textbf{A}}: who celebrates new year first in the world? \\
\textcolor{violet}{\textbf{B}}: Tonga and Kiritimati, part of Kiribati, are examples of the first places to welcome the New Year \\
\textcolor{violet}{\textbf{A}}: who celebrate new year last in the world? \\

\textcolor{violet}{\textbf{Final reply}}: \\
\textcolor{violet}{\textbf{B}}: Samoa and American Samoa are the last places to welcome the New Year, as they are the first to see the sunrise on January 1st. \\
\textcolor{violet}{\#\#\#}
\bigbreak
\textcolor{violet}{\textbf{Answer}}: 0.4
\bigbreak

\textcolor{violet}{\#\#\#} \\
\textcolor{violet}{\textbf{Dialog}}: \\
\textcolor{violet}{\textbf{A}}: Are there any other interesting aspects about Kanjani Eight's article? \\
\textcolor{violet}{\textbf{B}}: Kanjani Eito, stylized as Kanjani$\infty$) is a five-member Japanese boy band from Japan's Kansai region. \\
\textcolor{violet}{\textbf{A}}: When did the first album by Kanjani Eight come out? \\
\textcolor{violet}{\textbf{B}}: 03/15/2006 \\
\textcolor{violet}{\textbf{A}}: When was Kanjani Eight's first concert? \\
\textcolor{violet}{\textbf{B}}: December 2002 \\
\textcolor{violet}{\textbf{A}}: How many hit albums does Kanjani Eight have? \\
\textcolor{violet}{\textbf{B}}: 10 \\
\textcolor{violet}{\textbf{A}}: What songs are mentioned in the debut section? \\
\textcolor{violet}{\textbf{B}}: Naniwa Iroha Bushi“ \\
\textcolor{violet}{\textbf{A}}: Any other number one singles? \\
\textcolor{violet}{\textbf{B}}: Osaka Rainy Blues“ \\
\textcolor{violet}{\textbf{A}}: Any other interesting things I should know? \\
 
\textcolor{violet}{\textbf{Final reply}}: \\
\textcolor{violet}{\textbf{B}}: Not really. \\
\textcolor{violet}{\#\#\#}
 
\textcolor{violet}{\textbf{Answer}}: 0.8
\bigbreak
 
\textcolor{violet}{\#\#\#} \\
\textcolor{violet}{\textbf{Dialog}}: \\
\textcolor{violet}{\textbf{A}}: who becomes president if the president and vice president die in india \\
\textcolor{violet}{\textbf{B}}: An election to fill a vacancy in the office of President occurring  by reason of his death, resignation or removal, or otherwise shall be held as soon as possible after. \\
\textcolor{violet}{\textbf{A}}: how is the president elected in india \\
\textcolor{violet}{\textbf{B}}: The president is indirectly elected by an electoral college comprising the Parliament of India and the legislative assemblies of each of India's states and territories, who are all directly elected. \\
\textcolor{violet}{\textbf{A}}: when did india achieve independence \\
\textcolor{violet}{\textbf{B}}: India achieved independence from the British on 15 August 1947, initially as a dominion within the Commonwealth of Nations with George VI as king. \\
\textcolor{violet}{\textbf{A}}: what does India's constitution say about the president \\
 
\textcolor{violet}{\textbf{Final reply}}: \\
\textcolor{violet}{\textbf{B}}: The President of India (IAST: Bhārat Ganarājya Rāstrapatī) is the head of state of India and the commander-in-chief of the Indian Armed Forces. \\
\textcolor{violet}{\#\#\#}
 
\textcolor{violet}{\textbf{Answer}}: 1.0
\bigbreak
\end{minipage}}

\fbox{\begin{minipage}{50em}
\textcolor{violet}{\#\#\#} \\
\textcolor{violet}{\textbf{Dialog}}: \\
\textcolor{violet}{\textbf{A}}: How many members were in English rhythm and blues and rock band The Animals? \\
\textcolor{violet}{\textbf{B}}: The original line-up was Eric Burdon, Alan Price, Hilton Valentine, John Steel, and Bryan Chas Chandler. \\
\textcolor{violet}{\textbf{A}}: What happened at the reunion of the first incarnation of English rhythm and blues and rock band The Animals? \\
\textcolor{violet}{\textbf{B}}: They did a mini-tour in 1976 and shot a few videos of their new songs. \\
\textcolor{violet}{\textbf{A}}: Do you know any of the reunion of the first incarnation of English rhythm and blues and rock band The Animals' video titles by chance? \\
\textcolor{violet}{\textbf{B}}: They did a mini-tour in 1976 and shot a few videos of their new songs like Lonely Avenue and Please Send Me Someone to Love. \\
\textcolor{violet}{\textbf{A}}: Are all the members still alive and with English rhythm and blues and rock band The Animals? \\
 
\textcolor{violet}{\textbf{Final reply}}: \\
\textcolor{violet}{\textbf{B}}: The original band members are still alive, except for Chas Chandler, who died in 1996, and Bryan Chas Chandler, who died in 2006. \\
\textcolor{violet}{\#\#\#}
 
\textcolor{violet}{\textbf{Answer}}: 1.0
\bigbreak
 
\textcolor{violet}{\#\#\#} \\
\textcolor{violet}{\textbf{Dialog}}: \\
\textcolor{violet}{\textbf{A}}: when did ministry of corporate affairs issue ind as \\
\textcolor{violet}{\textbf{B}}: The Ministry of Corporate Affairs, in 2015, stipulated the adoption and applicability of IND AS. The MCA has since issued three Amendment Rules, one each in 2016, 2017, and 2018. \\
\textcolor{violet}{\textbf{A}}: what is the abbreviation IND AS \\
\textcolor{violet}{\textbf{B}}: Indian Accounting Standard (abbreviated as Ind-AS) is the Accounting standard adopted by companies in India. \\
\textcolor{violet}{\textbf{A}}: what preceded Ind AS \\
\textcolor{violet}{\textbf{B}}: India followed accounting standards from Indian Generally Acceptable Accounting Principle (IGAAP) prior to adoption of the Ind-AS \\
\textcolor{violet}{\textbf{A}}: are companies required to follow Ind AS \\
\textcolor{violet}{\textbf{B}}: Companies shall follow Ind AS either Voluntarily or Mandatorily. \\
\textcolor{violet}{\textbf{A}}: which companies is it mandatory for following Ind AS standards \\
 
\textcolor{violet}{\textbf{Final reply}}: \\
\textcolor{violet}{\textbf{B}}: It is mandatorily applied for such companies that are listed on the stock exchanges, companies with paid-up capital of more than five hundred crore rupees, companies with turnover of more than one thousand crore rupees, and companies with net worth of more than two thousand crore rupees. \\
\textcolor{violet}{\#\#\#} \\
 
\textcolor{violet}{\textbf{Answer}}: 1.0
\bigbreak
 
\textcolor{violet}{\#\#\#} \\
\textcolor{violet}{\textbf{Dialog}}: \\
\{context\} \\
\textcolor{violet}{\textbf{Final reply}}: \\
\{input\} \\
\textcolor{violet}{\#\#\#}
\end{minipage}}

\end{document}